\documentclass[11pt]{article}

\usepackage[final]{acl}

\usepackage{times}
\usepackage{latexsym}

\usepackage[T1]{fontenc}

\usepackage[utf8]{inputenc}

\usepackage{microtype}

\usepackage{inconsolata}

\usepackage{graphicx}

\usepackage{amsmath}
\usepackage{booktabs}
\usepackage{graphicx}
\usepackage{subcaption}
\usepackage{amsfonts}   
\usepackage{booktabs}
\usepackage{tabularx}

\title{SAMoRA: Semantic-Aware Mixture of LoRA Experts for Task-Adaptive Learning}

\author{
  \textbf{Boyan Shi\textsuperscript{1,3}},
  \textbf{Wei Chen\textsuperscript{2,$*$}},
  \textbf{Shuyuan Zhao\textsuperscript{1,3}},
  \textbf{Junfeng Shen\textsuperscript{1,3}},
  \\
  \textbf{Shengnan Guo\textsuperscript{1,3}},
  \textbf{Shaojiang Wang\textsuperscript{4,5,$*$}},
  \textbf{Huaiyu Wan\textsuperscript{1,3}}
  \\
  \textsuperscript{1}School of Computer Science and Technology, Beijing Jiaotong University, China \\
  \textsuperscript{2}Guangxi Key Lab of Trusted Software, Guilin University of Electronic Technology, China \\
  \textsuperscript{3}Beijing Key Lab of Traffic Data Mining and Embodied Intelligence, China \\
  \textsuperscript{4}Institute of AI for Industries, Chinese Academy of Sciences, China \\
  \textsuperscript{5}Nanjing Institute of Software Technology, China
  \\
  \small{
    \texttt{boyan118@bjtu.edu.cn}
  }
  \small{
    $^{\ast}$\textbf{Correspondence:} 
    \href{mailto:w_chen@guet.edu.cn}{\texttt{w\_chen@guet.edu.cn}}, 
    \href{mailto:wangshaojiang@iaii.ac.cn}{\texttt{wangshaojiang@iaii.ac.cn}}
  }
}

\begin{document}
\maketitle
\begin{abstract}
The combination of Mixture-of-Experts (MoE) and Low-Rank Adaptation (LoRA) has shown significant potential for enhancing the multi-task learning capabilities of Large Language Models. However, existing methods face two primary challenges: 
(1)Imprecise Routing in the current MoE-LoRA method fails to explicitly match input semantics with expert capabilities, leading to weak expert specialization.
(2)Uniform weight fusion strategies struggle to provide adaptive update strengths, overlooking the varying complexity of different tasks.
To address these limitations, we propose \textbf{SAMoRA} (\textbf{S}emantic-\textbf{A}ware \textbf{M}ixture \textbf{o}f Lo\textbf{RA} Experts), a novel parameter-efficient fine-tuning framework tailored for task-adaptive learning. Specifically, A \textbf{Semantic-Aware Router} is proposed to explicitly align textual semantics with the most suitable experts for precise routing. A \textbf{Task-Adaptive Scaling} mechanism is designed to regulate expert contributions based on specific task requirements dynamically. 
In addition, a novel regularization objective is proposed to jointly promote expert specialization and effective scaling. 
Extensive experiments on multiple multi-task benchmarks demonstrate that SAMoRA significantly outperforms the state-of-the-art methods and holds excellent task generalization capabilities. Code is available at \url{https://github.com/boyan-code/SAMoRA}
\end{abstract}

\section{Introduction}


Large Language Models (LLMs) have achieved impressive performance across a wide range of domains, particularly in natural language processing (NLP) tasks such as content generation and question answering~\cite{DBLP:journals/corr/abs-2507-01006, xu2023parameter, DBLP:conf/naacl/ChenHXWXC25}.
This success largely stems from their massive parameter counts and pre-training on large-scale, diverse corpora, which endow LLMs with strong generalization capabilities and robust performance across diverse and complex tasks~\cite{DBLP:conf/emnlp/QinZ0CYY23, DBLP:journals/jmlr/RaffelSRLNMZLL20, DBLP:conf/mm/ChenJWYCHX0025}, yet inevitably imposes a substantial parameter burden during fine-tuning.

\begin{figure}[t]
    \centering
    \begin{subfigure}[b]{1\linewidth}
        \centering
        \includegraphics[width=\linewidth]{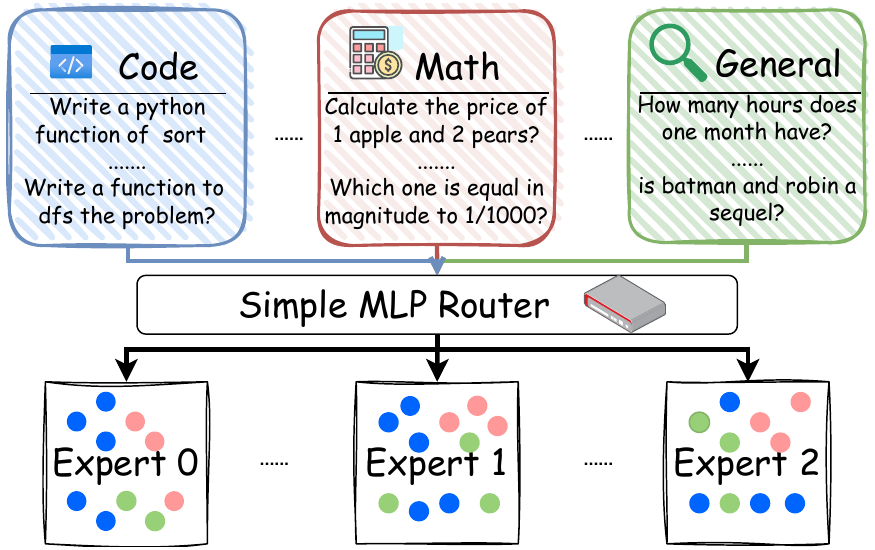}
        \caption{}
    \end{subfigure}
    
    \begin{subfigure}[b]{1\linewidth}
        \centering
        \includegraphics[width=\linewidth]{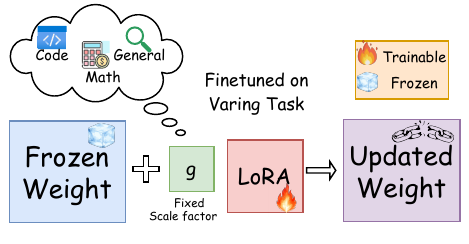}
        \caption{}
    \end{subfigure}
    
    \caption{
    \textbf{Illustration of limitations in existing mechanisms.}
    (a) MLP-based Routing: Fails to explicitly match tasks with expert capabilities, resulting in expert homogenization.
    (b) Uniform Weight Fusion: Applies a uniform update strength across diverse tasks, ignoring specific requirements and limiting multi-task generalization.
    }
    \label{fig:introduction_combined}
\end{figure}

To mitigate the computational burden of full fine-tuning, Low-Rank Adaptation (LoRA) has emerged as a leading Parameter-Efficient Fine-Tuning (PEFT) strategy~\cite{DBLP:conf/iclr/HuSWALWWC22}. 
LoRA injects trainable low-rank matrices into the frozen backbone and merges the updates via a uniform scaling factor. 
However, while effective for single tasks, this fixed structure limits performance in complex multi-task scenarios, as a single set of parameters cannot adequately handle diverse task requirements.
To address this, recent studies have integrated Mixture-of-Experts (MoE) architectures with LoRA (MoE-LoRA)~\cite{DBLP:conf/sigir/Liu00ZX0024}. 
These methods treat multiple LoRA modules as experts and employ a Multi-Layer Perceptron (MLP) based router to selectively activate them. 
While these approaches have demonstrated notable success in enhancing model capacity, they still face two critical challenges:

\textbf{(1) Current routing mechanisms fail to explicitly associate tasks with expert capabilities, leading to imprecise routing.} 
Existing MoE-LoRA methods rely on MLP routers that prioritize learned data distributions over actual expert proficiencies~\cite{DBLP:conf/nips/TianSG0024}. 
As illustrated in Figure~\ref{fig:introduction_combined}(a), these strategies fail to explicitly match input semantics with expert expertise, often resulting in homogenized experts that lack distinct roles. 
Consequently, this inability to specialize prevents the model from handling diverse requirements effectively, leading to suboptimal capabilities in multi-task scenarios.
\textbf{(2) Uniform weight fusion strategies fail to provide adaptive adjustments for diverse tasks, limiting multi-task generalization.} 
As shown in Figure~\ref{fig:introduction_combined}(b), standard approaches employ a globally fixed scale factor that applies a uniform update strength across all inputs. 
However, multi-task scenarios involve tasks with varying complexity, where some require significant parameter shifts while others need only minor adjustments. 
Applying a uniform strategy ignores these distinct requirements, forcing a rigid "one-size-fits-all" adaptation. 
This lack of flexibility prevents the model from effectively adapting to specific task needs, thereby constraining its overall generalization capability in complex multi-task environments.




To address these challenges, we propose \textbf{SAMoRA} (\textbf{S}emantic-\textbf{A}ware \textbf{M}ixture \textbf{o}f Lo\textbf{RA} Experts), a novel framework tailored for task-adaptive learning.
Specifically, SAMoRA consists of a  Semantic-Aware Router to explicitly align input semantics with expert capabilities, a Task-Adaptive Scaling mechanism to dynamically regulate expert contributions based on specific task demands, and specialized loss constraints to enforce expert distinctiveness and ensure robust multi-task performance.

The contributions of this work are as follows: 
\begin{itemize}
\item We propose \textbf{SAMoRA}, a novel MoE-LoRA framework enabling precise semantic-aware expert routing and significantly enhancing multi-task generalization capabilities.

\item We introduce a \textbf{Semantic-Aware Router} to enforce explicit alignment between input semantics and expert capabilities, coupled with a \textbf{Task-Adaptive Scaling} mechanism that dynamically regulates parameter updates to effectively adapt to diverse task requirements.

\item We design specialized loss constraints to enforce expert distinctiveness and regularize scaling factors, ensuring specialized expert roles and robust performance.
\item Extensive experiments across diverse multi-task benchmarks demonstrate that SAMoRA consistently outperforms existing baselines, achieving State-of-the-Art performance.
\end{itemize}

\section{Related Work}
\subsection{Mixture of Experts.}
MoE was initially proposed to decompose complex tasks into simpler subtasks, where a router dynamically assigns different inputs to specialized expert subnetworks~\cite{DBLP:journals/neco/JacobsJNH91}. A key later advancement was the sparsely-gated MoE, which activates only a small subset of experts per forward pass to significantly improve computational efficiency~\cite{DBLP:conf/iclr/ShazeerMMDLHD17}. This sparse-gating mechanism was subsequently extended to Transformer architectures, further enhancing training efficiency and model scalability~\cite{DBLP:conf/iclr/LepikhinLXCFHKS21}.
Subsequent strategies have further optimized routing mechanisms, such as simplified top-1 routing for stability~\cite{DBLP:journals/jmlr/FedusZS22} and differentiable soft routing for effective expert combination~\cite{DBLP:journals/tmlr/MuqeethLR24}. 

Despite these architectural improvements, current methods share a fundamental limitation: they rely on implicit routing strategies that lack explicit semantic guidance. These approaches typically map inputs to experts based on learned statistical distributions rather than establishing an explicit association between input semantics and expert capabilities. Consequently, the routing decision remains decoupled from actual expert specialization, hindering the model's ability to precisely match diverse inputs to the most suitable experts based on their semantic features.

\subsection{LoRA for Multi-Task Learning}
LoRA has attracted widespread attention due to its ability to achieve performance comparable to full fine-tuning under limited computational resources. 
However, its performance in complex multi-task scenarios remains suboptimal. 
To address this, several extensions have been proposed to enhance adaptability. 
MultiLoRA introduces a parallelized design with learnable scaling factors to decouple task-specific features~\cite{DBLP:journals/corr/abs-2311-11501}, while MTL-LoRA employs task-specific transformation matrices to capture distinct information~\cite{DBLP:conf/aaai/0001MSZLWSD0ZCT25}. 
Furthermore, methods like MoELoRA and HydraLoRA integrate MoE architectures, treating LoRA modules as experts to improve generalization and parameter efficiency~\cite{DBLP:conf/iclr/Liao0S0W25, DBLP:conf/nips/TianSG0024}.

Despite these architectural advancements, these methods share a fundamental limitation in their weight fusion mechanism. 
Most approaches rely on uniform scaling strategies to merge LoRA updates with the pre-trained model. 
This fixed approach ignores the varying complexity of different tasks, where some require significant parameter shifts while others need only minor adjustments. 
Consequently, applying the same update strength to all tasks fails to meet specific requirements, thereby limiting the model's overall multi-task adaptation performance.

\section{Preliminary}
\subsection{PEFT for LLMs}

PEFT for LLMs involves adapting pretrained models to downstream tasks by introducing a small set of trainable parameters $\Delta W$, while keeping the original model weights $W$ frozen. The model is jointly trained on multiple tasks in multi-task scenarios to learn shared and task-specific representations~\cite{DBLP:conf/iclr/WeiBZGYLDDL22}. The training objective is to fine-tune $\Delta W$ such that the conditional probability $P$ of autoregressively generating target sequences across all tasks is maximized. Formally, the training loss can be written as:
\begin{equation}
\begin{aligned}
\mathcal{L}_\text{task}(\Delta W)
= {} & \sum_{(s_{\text{in}}, s_{\text{out}}) \in \mathcal{D}}
      \sum_{i=1}^{|s_{\text{out}}|} \\
& \log P_{W + \Delta W}
\left(
s_{\text{out}}^{(i)}
\mid
s_{\text{in}}, s_{\text{out}}^{(<i)}
\right),
\label{eq:cross}
\end{aligned}
\end{equation}
where $\mathcal{D}$ denotes the training dataset containing input-output sentence pairs ($s_\text{in}, s_\text{out}$) from multiple tasks. This objective formalizes the autoregressive training process, where the model predicts target sentence incrementally by adapting only the incremental parameters $\Delta W$.

\subsection{Mixture of LoRA Experts}
LoRA implements PEFT by freezing the original pretrained weights $W$ and introducing two trainable low-rank matrices. For a weight matrix $W \in \mathbb{R}^{d_{\text{out}} \times d_{\text{in}}}$, these matrices are specifically defined as $A \in \mathbb{R}^{r \times d_{\text{in}}}$ and $B \in \mathbb{R}^{d_{\text{out}} \times r}$, where the rank $r$ is significantly smaller than the original dimensions. The resulting product $BA$  provides a low-rank update $\Delta W$ to $W$, enabling effective adaptation with minimal additional parameters~\cite{DBLP:conf/iclr/HuSWALWWC22}. The LoRA update process is illustrated in Figure~\ref{fig:introduction_combined}(b).

To leverage the parameter-efficiency of LoRA for complex multi-task scenarios, a promising direction in recent work has been to integrate it with MoE~\cite{DBLP:conf/aaai/0001MSZLWSD0ZCT25, DBLP:conf/icml/LiuWY0WCC24, DBLP:conf/coling/FengHZHW24}. By structuring multiple LoRAs as lightweight experts within the attention and feedforward layers of an LLM, the forward pass in such a layer is formalized as:
\begin{equation}
Y = W X + \sum_{i=1}^{N} g_i B_i A_i X,
\label{eq:pure_MoE_LoRA}
\end{equation}
where $X \in \text{Emb}(s_{\text{in}})$ is a hidden representation derived from the input sentence $s_{\text{in}}$, and $Y$ is the corresponding output. The set $\{A_i, B_i\}_{i=1}^N$ represents $N$ distinct LoRA experts. The gating weights $g_i$ are dynamically generated by a router conditioned on input $X$, determining which expert to activate.

\begin{figure*}[h]
    \centering
    \includegraphics[width=1\linewidth]{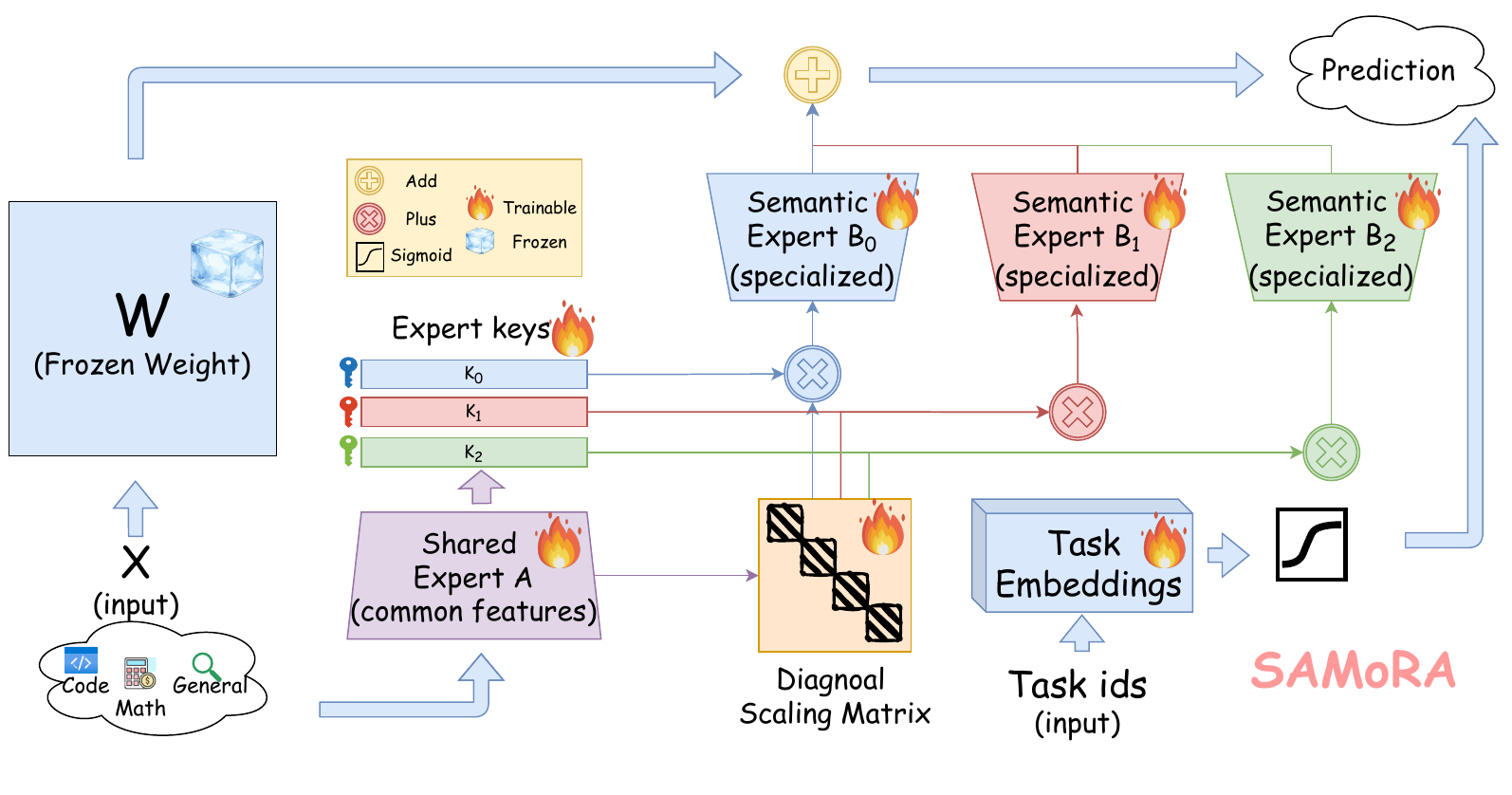}    
    \caption{Overview of our SAMoRA. We design a Semantic-Aware Router and a Task-Adaptive Scaling mechanism, integrated within an asymmetric MoE-LoRA architecture  consisting of a shared Expert A and multiple Semantic Experts B.}
    \label{fig:model}
\end{figure*}
\section{Methodology}
\label{sec:method}
As illustrated in Figure~\ref{fig:model}, SAMoRA integrates two core components: a \textbf{Semantic-Aware Router} designed to explicitly match input semantics with expert expertise, and a \textbf{Task-Adaptive Scaling} mechanism that dynamically regulates update strengths to meet specific task requirements. 
In the following, we introduce these components in detail.

\subsection{Semantic-Aware Router}
Most existing MoE approaches rely on MLP-based routing strategies that often fail to associate input contents with expert capabilities. 
To address this, we introduce a \textbf{Semantic-Aware Router} designed to explicitly match input semantics with expert expertise.

\paragraph{Semantic Extraction via Shared Expert.}
To realize explicit routing, the model must first effectively grasp the semantic intent of the input. Inspired by HydraLoRA~\cite{DBLP:conf/nips/TianSG0024}, we establish an asymmetric architecture by utilizing a single shared expert $A \in \mathbb{R}^{r \times d_{\text{in}}}$, while maintaining multiple experts $\{B_i\}_{i=1}^N$ to capture distinct semantic capabilities.
This shared component naturally functions as a semantic encoder, eliminating the need for a separate, decoupled routing network.
This shared module extracts a compact, unified semantic representation $\mathbf{h} = AX$ from the input $X$. By using the shared expert $A$, we ensure that the routing decision is grounded in the same feature space used for expert computation, facilitating consistent semantics aggregation. Building upon Eq.~\eqref{eq:pure_MoE_LoRA}, the core forward process is reformulated as:
\begin{equation}
Y = WX + \sum_{i=1}^N g_i B_i \mathbf{h} = WX + \sum_{i=1}^N g_i B_i (AX).
\label{eq:gate_SAMoRA}
\end{equation}

\paragraph{Explicit Matching with Expert Keys.}
With the extracted semantic features $\mathbf{h}$, the next step is to align them with the specific capabilities of the semantic experts $\{B_i\}_{i=1}^N$. To this end, we assign a trainable \textbf{Expert Key} $k_i \in \mathbb{R}^r$ to each expert $B_i$.
These keys function as semantic anchors, explicitly representing the unique specialization learned by each expert. During training, the keys are optimized alongside the experts, ensuring that $k_i$ moves closer to the semantic clusters that expert $B_i$ is best at handling. The routing score $g_i$ is then derived by measuring the Cosine Similarity between the input's semantic representation $\mathbf{h}$ and each expert key $k_i$:
\begin{equation}
g_{i} =
\frac{
\exp\left( \mathrm{cos}\left(\mathbf{h}, k_{i}\right) / \tau \right)
}{
\sum_{j=1}^N \exp\left( \mathrm{cos}\left(\mathbf{h}, k_{j}\right) / \tau \right)
}, 
\end{equation}
where $\tau$ is a temperature coefficient that regulates the strictness of the matching. A smaller $\tau$ sharpens the distribution, forcing the router to strictly select only the expert with the highest semantic alignment, while a larger $\tau$ softens this constraint, allowing for broader expert collaboration. This mechanism ensures that inputs are routed based on explicit semantic similarity rather than implicit statistical bias.

\subsection{Task-Adaptive Scaling}
As illustrated in Figure~\ref{fig:introduction_combined}(b), standard LoRA employs a uniform scaling factor to merge the updates. 
However, this fixed approach is problematic in multi-task scenarios as it ignores the varying complexity of different tasks. 
Some tasks require significant parameter shifts while others need only minor adjustments. 
Consequently, applying the same update strength to all tasks fails to meet specific requirements, limiting the model's adaptability. To address this, we propose a \textbf{Task-Adaptive Scaling} mechanism that dynamically regulates the update magnitude based on specific task demands.

\paragraph{Spectral Initialization via SVD.}
First, inspired by recent work~\cite{yuan2025moore, DBLP:conf/aaai/Zhao0SZLW25}, we aim to ensure our asymmetric structure starts with a theoretically grounded scale alignment. We introduce a trainable Diagonal Scaling Matrix $S \in \mathbb{R}^{r \times r}$ positioned between the shared expert $A$ and the semantic experts $B_i$.
By performing Singular Value Decomposition (SVD) on the pre-trained weight $W = U \Sigma V^\top$, we initialize $S$ using the top-$r$ singular values:
\begin{equation}
    S = \Sigma_{1:r, 1:r} = \mathrm{diag}(\sigma_1, \ldots, \sigma_r).
\end{equation}
This design aligns our components with the dominant semantic directions of the original weights, providing a stable structural basis for subsequent adaptation.

\paragraph{Task-Dependent Dynamic Regulation.}
Building upon this aligned basis, we introduce a task-driven mechanism to dynamically control the fusion ratio. We assign a learnable \textbf{Task Embedding} $e_{\text{task}} \in \mathbb{R}^{d_g}$ to each task, which captures latent task characteristics such as complexity and domain divergence.
To determine the optimal update strength for a given task, we project this embedding into a scalar gating factor $g_{\text{task}} \in (0, 1)$ via a non-linear mapping:
\begin{equation}
    g_{\text{task}} = \sigma(W_{\text{gate}} e_{\text{task}} + b_{\text{gate}}),
\end{equation}
where $\sigma(\cdot)$ is the sigmoid function. This mechanism allows the model to dynamically adjust the update strength based on input features. It assigns larger scales for tasks needing significant adaptation and smaller scales for those requiring only minor adjustments, effectively meeting diverse task requirements.

By integrating the SVD-based alignment and task-dependent regulation into the formulations of Eq.~\eqref{eq:pure_MoE_LoRA} and Eq.~\eqref{eq:gate_SAMoRA}, the final output $Y$ is derived as:
\begin{equation}
    Y = WX + g_{\text{task}} \sum_{i=1}^N g_i B_i (S A X).
\end{equation}

\subsection{Training Objective}
\label{sec:objective}
To ensure the effective implementation of our proposed mechanisms, we incorporate two specialized regularization terms alongside the standard LLM generation loss. 
Specifically, these terms are designed to align Expert Keys with their corresponding experts and impose the necessary SVD constraints for the Task-Adaptive Scaling mechanism. 
The total training objective is formulated as:
\begin{equation}
\mathcal{L}_\mathrm{total} = \mathcal{L}_\mathrm{task} + \lambda_\mathrm{orth} \cdot \mathcal{L}_\mathrm{orth} + \lambda_\mathrm{match} \cdot \mathcal{L}_\mathrm{match},
\end{equation}
where $\mathcal{L}_\mathrm{task}$ denotes the multi-task language modeling loss Ep. ~\eqref{eq:cross}, and $\lambda_\mathrm{orth}$, $\lambda_\mathrm{match}$ are scalar hyperparameters weighting the auxiliary constraints.

\begin{table*}[t]
\centering
\resizebox{\textwidth}{!}{
\begin{tabular}{lc ccccccccc c}
\toprule
\textbf{Method} & \textbf{TP (\%)} & \textbf{BoolQ} & \textbf{PIQA} & \textbf{SIQA} & \textbf{WinoG} & \textbf{ARC-C} & \textbf{ARC-E} & \textbf{OBQA} & \textbf{HellaS} & \textbf{CSQA} & \textbf{Avg.} \\
\midrule
\multicolumn{11}{c}{\textit{Backbone: Llama3.1-8B}} \\
\midrule
LoRA$^{\dag}$  & 2.09 & 70.43 & 82.97 & 76.00 & 71.11 & 77.56 & 85.77 & 81.60 & 93.00 & 77.40 & 79.54 \\
MultiLoRA & 0.26 & 70.95 & 80.81 & 80.91 & 82.15 & 71.70 & 86.12 & 80.60 & 94.01 & 80.34 & 80.84 \\
MoORE & 0.77 & \underline{74.49} & 88.63 & \underline{82.99} & 87.74 & 79.95 & 88.80 & 86.20 & \underline{95.48}  & \underline{84.60} & 85.43 \\
HydraLoRA & 0.17 & 74.31 & \underline{90.15} & 82.49 & \underline{88.47} & 84.06 & 92.18 & 87.80 & 93.18  & 83.81 & 86.27 \\
MTL-LoRA & 0.16 & 74.34 & 89.90 & 82.95 & 88.08 & \underline{84.55} & \underline{93.81} & \underline{88.20} & 95.15  & 83.94 & \underline{86.77} \\
\textbf{SAMoRA (Ours)} & 0.15 & \textbf{74.89} & \textbf{90.37} & \textbf{83.32} & \textbf{88.95} & \textbf{86.35} & \textbf{94.70} & \textbf{89.80} & \textbf{95.97}  & \textbf{84.85} & \textbf{87.64} \\
\midrule
\multicolumn{11}{c}{\textit{Backbone: Qwen3-8B}} \\
\midrule
LoRA & 0.74 & 73.80 & \underline{91.45} & 83.00 & 88.39 & 92.40 & 97.60 & 90.20 & 94.60  & 86.32 & 88.64 \\
MultiLoRA & 0.29 & 71.89 & 89.88 & 81.83 & 83.89 & 92.15 & \underline{97.60} & 90.60 & 93.07  & 85.74 & 87.64 \\
MoELoRA & 0.56 & \underline{73.90} & 91.18 & 81.47 & 83.10 & 92.49 & 97.34 & 89.60 & 92.30  & 84.43 & 87.31 \\
HydraLoRA & 0.16 & 73.14 & 90.69 & \underline{83.21} & 87.92 & \textbf{92.90} & 97.47 & 89.40 & 94.60  & \underline{87.01} & 90.33 \\
MoORE & 0.84 & 73.60 & 91.26 & 80.80 & 86.55 & 90.10 & 93.30 & 90.20 & 94.09  & 86.56 & 90.28 \\
MTL-LoRA & 0.16 & 73.51 & 91.13 & 82.08 & \underline{88.87} & 92.15 & 97.55 & \underline{91.40} & \underline{95.47}  & 86.08 & \underline{90.98} \\
\textbf{SAMoRA (Ours)} & 0.15 & \textbf{74.68} & \textbf{92.00} & \textbf{83.78} & \textbf{88.95} & \underline{92.58} & \textbf{97.94} & \textbf{91.80} & \textbf{96.01}  & \textbf{87.31} & \textbf{91.71} \\
\bottomrule
\end{tabular}
}
\caption{Results of comparison experiments across Commonsense Reasoning benchmarks. TP indicates Trainable Parameters (\%). $^{\dag}$means the results from MoORE~\cite{yuan2025moore}. \textbf{Bold}: Best results; \underline{Underline}: Second-best results.}
\label{tab:commonsense}
\end{table*}

\begin{table*}[t]
\centering
\setlength{\tabcolsep}{5pt}
\begin{tabular}{lc ccccccc c}
\toprule
\textbf{Model} & \textbf{TP (\%)} & \textbf{CoLA} & \textbf{MNLI} & \textbf{MRPC} & \textbf{QNLI} & \textbf{QQP} & \textbf{RTE} & \textbf{SST2} & \textbf{Avg.} \\
 & & \small{(Mcc.)} & \small{(Acc.)} & \small{(Acc.)} & \small{(Acc.)} & \small{(Acc.)} & \small{(Acc.)} & \small{(Acc.)} & \\
\midrule
\multicolumn{10}{c}{\textit{Backbone: Qwen3-8B}} \\
\midrule
LoRA & 0.21 & 64.06 & 91.84 & 88.20 & \underline{96.01} & 91.12 & 91.16 & 96.50 & 88.41 \\
MultiLoRA & 0.67 & 58.50 & 90.83 & 80.88 & 94.12 & 89.57 & 85.56 & \underline{97.02} & 85.21 \\
MoeLoRA & 0.60 & 67.01 & 91.71 & 83.82 & 95.88 & 90.07 & 91.69 & 96.67 & 88.12 \\
HydraLoRA & 0.20 & \underline{67.04} & 91.90 & 85.04 & 90.51 & 90.68 & 75.45 & 96.55 & 85.31 \\
MTL-LoRA & 0.20 & 66.32 & \underline{91.93} & \textbf{89.46} & 95.77 & \underline{91.39} & \underline{92.77} & 96.67 & \underline{89.18} \\
\midrule
\textbf{SAMoRA (Ours)} & 0.18 & \textbf{69.75} & \textbf{91.96} & \underline{89.21} & \textbf{96.17} & \textbf{91.41} & \textbf{94.22} & \textbf{97.13} & \textbf{89.98} \\
\hspace{1em} w/o Router & 0.20 & 68.19 & 92.08 & 89.46 & 95.91 & 91.41 & 91.33 & 97.13 & 89.36 \\
\hspace{1em} w/o Scaling & 0.18 & 66.43 & 91.93 & 88.97 & 95.93 & 90.84 & 91.33 & 96.90 & 88.90 \\
\hspace{1em} w/o $\mathcal{L}_\text{orth} $ & 0.15 & 68.32 & 91.99 & 87.99 & 96.11 & 90.89 & 90.61 & 97.01 & 88.99 \\
\hspace{1em} w/o $\mathcal{L}_\text{match}$ & 0.15 & 68.73 & 91.88 & 87.25 & 95.97 & 90.63 & 91.69 & 97.01 & 89.02 \\
\bottomrule
\end{tabular}
\caption{Results of comparison experiments across GLUE benchmark. The upper block presents the baselines, while the lower block reports the performance of SAMoRA and its ablation variants. \textbf{Bold}: Best results; \underline{Underline}: Second-best results.}
\label{tab:glue}
\end{table*}

\paragraph{Orthogonality Regularization for Scale Decoupling.}
We introduce an orthogonality regularization term $\mathcal{L}_\mathrm{orth}$ to strictly decouple \textit{directional semantics} from \textit{magnitude scaling}.
In our SVD-based design, the diagonal matrix $S$ and the gating factor $g_{\text{task}}$ are intended to handle all "scaling" effects.
Specifically, we force the rows of the shared encoder $A$ and the columns of each semantic expert $B_i$ to be orthonormal:
\begin{equation}
\mathcal{L}_\mathrm{orth} = \| AA^\top - I \|_F^2 + \sum_{i=1}^N \| B_i^\top B_i - I \|_F^2,
\end{equation}
where $I \in \mathbb{R}^{r \times r}$ is the identity matrix. By enforcing this constraint, $A$ and $B_i$ focus purely on learning distinct semantic directions, ensuring that the control of adaptation strength remains exclusively within the purview of our Task-Adaptive Scaling mechanism.

\paragraph{Semantic Match Regularization via KL Divergence.}
The effectiveness of our Semantic-Aware Router hinges on the semantic consistency between the learnable key $k_i$ and the functional specialization of the expert $B_i$. 
Any misalignment between $k_i$ and $B_i$ inevitably leads to erroneous expert selection. 
To mitigate this, we introduce a regularization loss that explicitly minimizes the divergence between $k_i$ and the semantic representation derived from $B_i$. We detail the specific implementation steps as follows.

\textit{(1) Extracting Representative Vectors.} Since the expert $B_i \in \mathbb{R}^{d_\text{out} \times r}$ is a matrix while the key $k_i \in \mathbb{R}^r$ is a vector, we obtain a representative vector $b_i$ from each expert. This is achieved by mean-pooling the row vectors of $B_i$, which aggregates the features learned by that expert:
\begin{equation}
    b_i = \frac{1}{d_\text{out}} \sum_{j=1}^{d_\text{out}} B_i^{(j)} \in \mathbb{R}^r.
\end{equation}

\textit{(2) Alignment via Distribution Matching.} To align the routing key with the expert's actual capability, we map both the key $k_i$ and the semantic centroid $b_i$ into probability distributions ($P_{k}^{(i)}$ and $P_{b}^{(i)}$) via Softmax. We then minimize the Kullback-Leibler (KL) divergence between them:
\begin{equation}
\mathcal{L}_\mathrm{match} = \frac{1}{N} \sum_{i=1}^{N} D_\mathrm{KL}\left(P_{b}^{(i)} \parallel P_{k}^{(i)}\right).
\end{equation}
Crucially, we employ the direction $D_\mathrm{KL}(P_{\text{Expert}} \parallel P_{\text{Key}})$. This effectively treats the expert's functional distribution $P_{b}^{(i)}$ as the target, compelling the key $P_{k}^{(i)}$ to shift towards and accurately represent the expert's specialization. This ensures consistency between the routing keys and the actual expert characteristics.

\subsection{Complexity Analysis}
\label{sec:complexity}

To demonstrate the computational efficiency and parameter economy of our framework, we compare the complexity of SAMoRA with the standard MoE-LoRA paradigm. For a comprehensive breakdown of all baselines and the detailed analysis process, please refer to Appendix~\ref{app:complexity}.

Standard MoE-LoRA architectures typically assign independent down-projection and up-projection matrices to each expert. This results in a parameter complexity of $\mathcal{O}(N(d_\text{in} + d_\text{out})r)$ and necessitates high-dimensional computations for routing, incurring a cost of $\mathcal{O}(N d_\text{in})$.

In contrast, SAMoRA optimizes both storage and inference efficiency through its asymmetric design and low-rank routing mechanism. 
Specifically, by using a shared expert $A$, SAMoRA eliminates the redundancy of learning separate input projections, reducing the parameter complexity to $\mathcal{O}((d_\text{in} + N d_\text{out})r)$. 
Furthermore, unlike standard methods that calculate routing scores in the high-dimensional input space ($d_\text{in}$), SAMoRA performs routing in the low-rank latent space ($r$). 
Given that $r \ll d_\text{in}$, this design significantly reduces the routing FLOPs from $\mathcal{O}(N d_\text{in})$ to $\mathcal{O}(Nr)$, ensuring minimal latency overhead during inference.

Overall, SAMoRA achieves a substantial reduction in both parameter count and computational cost compared to other MoE-LoRA baselines, offering a superior trade-off between model capacity and efficiency.

\section{Experiments}
\subsection{Experiment Setting}
\paragraph{Dataset}
We evaluate SAMoRA on two challenging multi-task benchmarks that target different capabilities of LLMs:
\textbf{(1) Commonsense Reasoning}: A curated benchmark comprising nine representative commonsense reasoning tasks: ARC-Challenge (ARC-C), ARC-Easy (ARC-E)~\cite{DBLP:journals/corr/abs-1803-05457}, OpenBookQA (OBQA)~\cite{DBLP:journals/corr/abs-1809-02789}, PIQA~\cite{DBLP:conf/aaai/BiskZLGC20}, SocialIQA (SIQA)~\cite{DBLP:journals/corr/abs-1904-09728}, BoolQ~\cite{DBLP:conf/nips/WangPNSMHLB19}, HellaSwag (HellaS)~\cite{DBLP:conf/acl/ZellersHBFC19}, Winogrande (WinoG)~\cite{DBLP:journals/cacm/SakaguchiBBC21} and CommonsenseQA(CSQA)~\cite{talmor2019commonsenseqa}. These datasets cover diverse commonsense challenges, including science QA, physical and social reasoning, and everyday inference, and are widely used to evaluate the multi-task capabilities of LLMs. \textbf{(2) Natural Language Understanding}: We use widely used subset of seven tasks from the GLUE benchmark~\cite{DBLP:conf/iclr/WangSMHLB19}, including CoLA, SST-2, MRPC, QQP, MNLI, QNLI, and RTE. These tasks assess linguistic phenomena such as grammaticality, sentiment analysis, paraphrase detection, and textual entailment, thus comprehensively evaluating general language understanding capabilities. 

Following the same train-test split protocol and instruction prompts as in prior works~\cite{DBLP:conf/aaai/0001MSZLWSD0ZCT25, DBLP:conf/icml/LiuWY0WCC24}, we conduct our evaluation. Detailed descriptions of the data splits and prompt formats are provided in Appendix~\ref{app:datasets}.

\paragraph{Implementation Details.}
We conduct experiments using Qwen3-8B~\cite{DBLP:journals/corr/abs-2505-09388} and LLaMA3.1-8B~\cite{DBLP:journals/corr/abs-2407-21783} as the backbone architectures. We compare SAMoRA against a comprehensive set of competitive baselines, including LoRA~\cite{DBLP:conf/iclr/HuSWALWWC22}, MultiLoRA~\cite{DBLP:journals/corr/abs-2311-11501}, MoELoRA~\cite{DBLP:conf/sigir/Liu00ZX0024}, HydraLoRA~\cite{DBLP:conf/nips/TianSG0024}, MTL-LoRA~\cite{DBLP:conf/aaai/0001MSZLWSD0ZCT25}, and MoORE~\cite{yuan2025moore}.
To ensure a fair comparison, we modify the hyperparameters of the baselines to make the number of trainable parameters comparable for each method.
We report detailed training settings for all baselines in Appendix~\ref{app:details}.

\subsection{Overall Performance}

As shown in Table~\ref{tab:commonsense} and Table~\ref{tab:glue}, SAMoRA consistently outperforms existing baselines on both Llama3.1-8b and Qwen3-8b across Commonsense Reasoning and GLUE benchmarks, while maintaining strong parameter efficiency. 
Compared to the single-adapter method LoRA, SAMoRA demonstrates clear advantages in handling diverse tasks, underscoring the importance of multi-expert architectures in multi-task adaptation.

Compared with MTL-LoRA and HydraLoRA, which rely on conventional MLP-based routers, SAMoRA enables more accurate and flexible expert selection through its semantic-aware routing mechanism. 
Regarding MoORE, it attempts to leverage the original LLM weights by exclusively training the router. However, this approach performs poorly due to the limited number of trainable parameters, which proves insufficient for effective adaptation on downstream tasks.
Furthermore, while MoELoRA introduces task-specific experts and MultiLoRA assigns a separate trainable scale factor to each LoRA module, they fail to account for task-specific characteristics and varying task complexity simultaneously. 
In contrast, SAMoRA introduces a task-adaptive scaling mechanism that dynamically modulates this balance, enabling more precise and efficient adaptation across diverse tasks with fewer trainable parameters.

\subsection{Ablation Study}

To better understand the effectiveness of each component in SAMoRA, we conduct a comprehensive ablation study. We evaluate the impact of the proposed semantic-aware router by comparing it against a conventional MLP-based router ($w/o$ Router). We assess the contribution of the task-adaptive scaling mechanism by removing it across all tasks($w/o$ Scaling). In addition, we examine the influence of the auxiliary losses by removing the orthogonality loss ($ w/o$ $ \mathcal{L}_\text{orth}$) and semantic match loss ($ w/o$ $ \mathcal{L}_\text{match}$), respectively. The results are summarized in Table~\ref{tab:glue}, and further implementation details are provided in Appendix~\ref{app:c1}.

As presented in Table~\ref{tab:glue}, SAMoRA consistently achieves the best performance across all tasks, validating the synergy of its components. Notably, removing the task-adaptive scaling mechanism ($w/o$ Scaling) leads to the most significant performance degradation (a sharp drop from 69.75\% to 66.43\% on CoLA), underscoring its critical role in resolving task conflicts and mitigating negative transfer. Similarly, replacing the semantic-aware router with a standard MLP ($w/o$ Router) results in a clear decline, confirming the necessity of explicit semantic alignment for precise expert allocation. Furthermore, excluding the auxiliary regularization terms ($w/o$ $\mathcal{L}_\text{orth}$ and $w/o$ $\mathcal{L}_\text{match}$) also impairs overall results, demonstrating their importance in maintaining expert distinctiveness and stabilizing training.

\begin{figure}[h]
    \centering
    \includegraphics[width=1\linewidth]{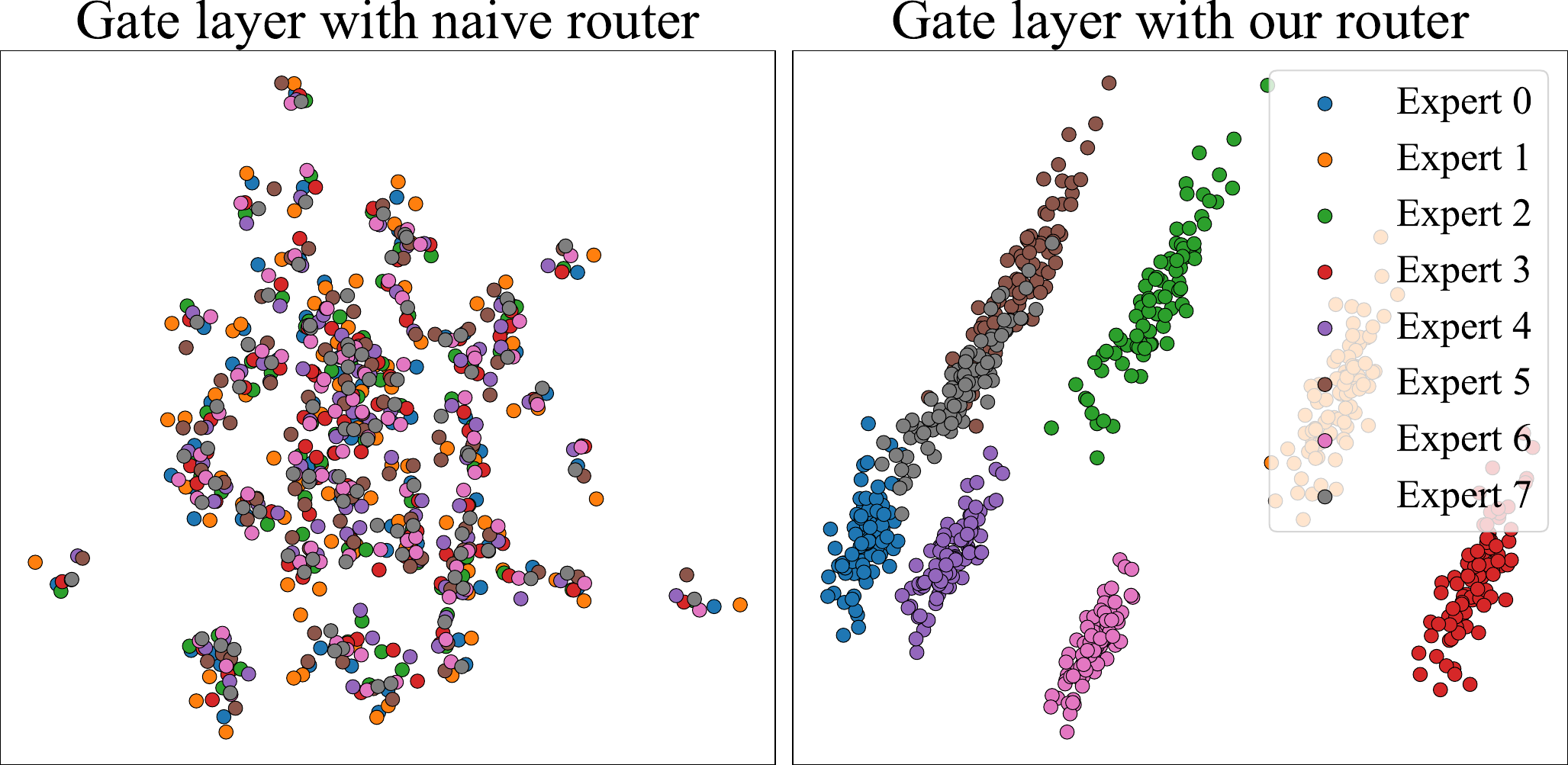} 
    \caption{PCA visualization of expert features extracted from the final gate layer trained on Commonsense Reasoning dataset.}
    \label{fig:pca}
\end{figure}

\paragraph{Analysis of Semantic-Aware Router.}
To investigate expert specialization, we visualize the PCA projections of the latent representations derived from the Semantic Expert $B$ matrices. 
As illustrated in Figure~\ref{fig:pca}, the standard MLP router results in entangled clusters with blurred boundaries. 
In contrast, our Semantic-Aware Router yields distinct and well-separated clusters for the Semantic Expert $B$ modules, explicitly confirming that each expert has specialized in a specific semantic subspace. 
Detailed experimental settings are provided in Appendix~\ref{app:c2}.

\begin{figure}[h]
    \centering
    \includegraphics[width=1\linewidth]{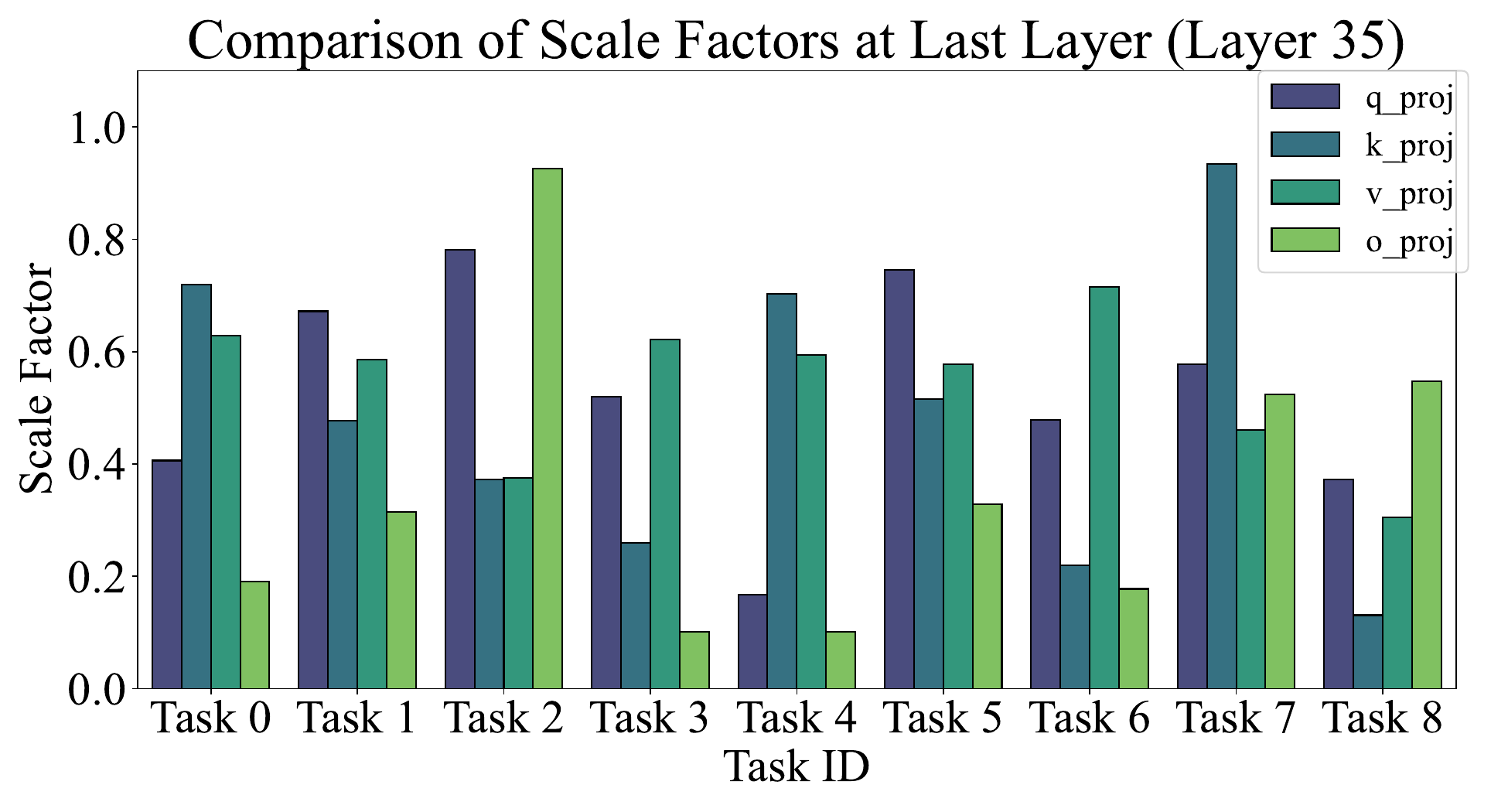}
    \caption{Visualization of task scaling factors across tasks trained on the Commonsense Reasoning dataset.}
    \label{fig:placeholder}
\end{figure}

\paragraph{Analysis of Task-Adaptive Scaling.} 
To validate the effectiveness of our mechanism, we visualize the learned scaling factors of SAMoRA trained on Qwen3-8B. 
Figure~\ref{fig:placeholder} displays the factors for the query ($q$), key ($k$), value ($v$), and output ($o$) projections within the final attention layer. 
The observed variations across different tasks validate the effectiveness of our proposed mechanism. 

\subsection{Sensitivity Analysis}
We conduct a comprehensive sensitivity analysis on key hyperparameters, including model architecture ($N, r, d_g$) and training objectives ($\lambda_\mathrm{orth}, \lambda_\mathrm{KL}, \tau$), with detailed results provided in Appendix~\ref{app:d}. 
Overall, the model exhibits strong robustness across varying configurations. 
Notably, regarding the task embedding dimension $d_g$, we observe that compact embeddings are sufficient for effective routing; increasing $d_g$ to excessive levels introduces unnecessary complexity that hinders convergence. 


\section{Conclusion}
In this paper, we propose SAMoRA, a novel PEFT framework significantly enhancing multi-task generalization. 
By ensuring precise expert routing and dynamic task adaptation, our approach effectively secures robust and superior performance across diverse multi-task scenarios.
Extensive experiments demonstrate that SAMoRA consistently outperforms existing baselines, achieving a favorable trade-off between performance and parameter efficiency.

\section*{Limitations}

In this paper, we conduct experiments on Commonsense Reasoning and GLUE benchmarks by fine-tuning models at the 8B parameter scale. 
Due to limited computational resources, the scalability of our framework to significantly larger foundation models (e.g., 70B scale or above) has not yet been empirically verified. 
Furthermore, there is a broader range of application scenarios unexplored, particularly in the multimodal domain, such as visual instruction tuning and visual question answering tasks. 
We plan to extend our method to these large-scale and multimodal settings in future work to further explore its generalization capabilities.

\section*{Acknowledgments}

This work was suported by Frontier Technologies R\&D Program of Jiangsu (Grant No. BF2024052), Nanjing Municipal Science and Technology Bureau (Grant No.202512136), and Chengdu Science and Technology Program (Grant No.2025-YF08-00097-GX).

\bibliography{custom}

\appendix

\section{Complexity Analysis}
\label{app:complexity}
\begin{table*}[t]
    \centering
    \renewcommand{\arraystretch}{1.3}
    \setlength{\tabcolsep}{6pt}
    \begin{tabular}{l c c}
        \toprule
        \textbf{Method} & \textbf{\# Learnable Parameters} & \textbf{Computational Complexity} \\
        \midrule
        LoRA & $(d_\text{in} + d_\text{out})r$ & $\mathcal{O}((d_\text{in} + d_\text{out})r)$ \\
        MoELoRA & $N(d_{in} + d_{out})r + Kd_g + Nd_g$ & $\mathcal{O}(N(d_{in} + d_{out})r + d_g)$ \\
        MTL-LoRA & $(d_\text{in} + Nd_\text{out})r + K r$ & $\mathcal{O}( r(1+ d_\text{in} + Nd_\text{out}))$ \\
        HydraLoRA & $(d_\text{in} + N d_\text{out})r + N d_\text{in}$ & $\mathcal{O}(d_\text{in} + (d_\text{in} + N d_\text{out})r)$ \\
        \midrule
        \textbf{SAMoRA (Ours)} & $(d_\text{in} + N + N d_\text{out})r + Kd_g + d_g$ & $\mathcal{O}((d_\text{in} + N + Nd_\text{out})r + d_g)$ \\
        \bottomrule
    \end{tabular}
    \caption{Comparison of learnable parameters and computational complexity. Notations: $d_\text{in}/d_\text{out}$ are input/output dimensions, $r$ is the rank, $N$ is the expert number, $K$ is the task number. $d_g$ denote task embedding sizes for MoELoRA and SAMoRA. SAMoRA achieves a superior trade-off by combining asymmetric experts with efficient routing.}
    \label{tab:complexity_comparison}
\end{table*}
\subsection{Theoretical Analysis}

In this section, we analyze the theoretical complexity of SAMoRA in terms of trainable parameters and computational overhead. We compare our method against standard LoRA~\cite{DBLP:conf/iclr/HuSWALWWC22} and representative MoE-based PEFT frameworks, including MoeLoRA~\cite{DBLP:conf/sigir/Liu00ZX0024}, HydraLoRA~\cite{DBLP:conf/nips/TianSG0024} and MTL-LoRA~\cite{DBLP:conf/aaai/0001MSZLWSD0ZCT25}.

For clarity, we define the following notations: $d_\text{in}$ and $d_\text{out}$ denote the input and output dimensions of the adapter layer, respectively. $r$ represents the low-rank dimension, $N$ is the number of experts, and $K$ is the number of tasks.

\paragraph{Parameter Efficiency.}
The comparison of learnable parameters is summarized in Table~\ref{tab:complexity_comparison}.
\begin{itemize}
    \item \textbf{Standard LoRA} employs a single pair of low-rank matrices per layer, resulting in $(d_\text{in} + d_\text{out})r$ parameters. It serves as the most parameter-efficient baseline but lacks multi-task flexibility.
    
    \item \textbf{MoELoRA} adopts a task-conditioned routing mechanism. It introduces a task embedding layer ($K d_g$) and a router projection matrix ($d_g N$) to generate routing probabilities based on task IDs. Unlike the asymmetric design in SAMoRA, MoELoRA maintains \textit{fully independent} low-rank experts. Consequently, its parameter complexity for the adapters is $N(d_\text{in} + d_\text{out})r$, which is significantly higher than shared-weight approaches. The total parameter count is given by $N(d_\text{in} + d_\text{out})r + K d_g + Nd_g$, where $d_g$ is the task embedding dimension.
    
    \item \textbf{HydraLoRA}, \textbf{MTL-LoRA} and \textbf{SAMoRA} adopt an \textit{asymmetric expert architecture}. To optimize parameter efficiency, we share the projection matrix on the input side ($A \in \mathbb{R}^{d_\text{in} \times r}$), while maintaining $N$ expert-specific matrices on the output side ($B \in \mathbb{R}^{r \times d_\text{out}}$). This design reduces the complexity from the standard MoE's $N(d_\text{in} + d_\text{out})r$ to $(d_\text{in} + N d_\text{out})r$.

    \item \textbf{MTL-LoRA} creates task-specific experts, scaling the number of parameters linearly with the number of tasks $K$. This results in a significantly higher parameter count of approximately $K N (d_\text{in} + d_\text{out})r$, making it less scalable for scenarios with many tasks.
\end{itemize}

\paragraph{Computational Overhead.}
Our SAMoRA framework introduces minimal computational overhead. The \textbf{Semantic-Aware Router} requires a lightweight projection from $d_\text{in}$ to the rank space $r$ (where $r \ll min( d_\text{in}, d_\text{out})$), adding only $\mathcal{O}(Nr)$ operations. The \textbf{Task-Adaptive Scaling} mechanism introduces a lightweight parameter set of size $K  d_g$ to capture task-specific characteristics. Since the scaling process primarily involves element-wise multiplications, the resulting computational overhead is negligible compared to the matrix multiplications in the backbone model.

\section{Experimental Setup}
\label{app:setup}

\subsection{Datasets and Prompts}
\label{app:datasets}
Following the experimental setup in~\cite{DBLP:conf/aaai/0001MSZLWSD0ZCT25}, we summarize the statistics for the Commonsense Reasoning and GLUE benchmarks in Table~\ref{tab:com_detail} and~\ref{tab:glue_detail}, respectively. 
The corresponding prompt templates used are detailed in  Table~\ref{tab:nlu_prompts}.

\begin{table}[ht]
\centering
\begin{tabular}{l|rrr}
\toprule
\textbf{Corpus} & {\textbf{\#Train}} & \textbf{\#Val.} & \textbf{Metrics}\\
\midrule
BoolQ              & 9427                 & 3270 & Accuarcy\\
PIQA               & 16100                & 1840 & Accuarcy \\
SocialIQA          & 33410                & 1954 & Accuarcy \\
WinoGrande         & 9248                 & 1267 & Accuarcy \\
ARC-Challenge      & 1119                 & 1172 & Accuarcy\\
ARC-Easy           & 2250                 & 2380 & Accuarcy\\
OpenBookQA         & 4957                 & 500  & Accuarcy\\
HellaSwag          & 39905                & 10042 & Accuarcy\\
CommonsenseQA      & 9741                 & 1140 & Accuarcy\\
\bottomrule
\end{tabular}
\label{tab:commonsense_tasks}
\caption{The basic information of Commonsense Reasoning Dataset}
\label{tab:com_detail}
\end{table}

\begin{table}[ht]
\centering
\resizebox{0.5\textwidth}{!}{
\begin{tabular}{l|rrrll}
\toprule
\textbf{Corpus} & \textbf{\#Train} & \textbf{\#Validation} & \textbf{Metrics}\\
\midrule
CoLA & 8,551 & 1,043 & Matthew's Corr.\\
MNLI & 392,702 & 9,815 & Accuracy\\
MRPC & 3,668 & 408 & Accuracy\\
QNLI & 104,743 & 5,463 & Accuracy\\
QQP & 363,846 & 40,430 & Accuracy\\
RTE & 2,490 & 277 & Accuracy\\
SST-2 & 67,349 & 872 & Accuracy\\
\bottomrule
\end{tabular}
}
\caption{The basic information of GLUE Benchmark}
\label{tab:glue_detail}
\end{table}

\subsection{Implementation Details}
\label{app:details}
We implement all methods using the PyTorch framework. 
Detailed hyperparameter configurations for our proposed SAMoRA and all baseline methods are summarized in Table~\ref{tab:hyperparameters}. 

\section{Extended Analyses and Ablation Studies}
\label{app:analysis}

\subsection{Setup of Ablation Variants}
\label{app:c1}
To rigorously evaluate the contribution of each component in SAMoRA, we conduct ablation studies using the Qwen3-8B model on the GLUE benchmark. The specific configurations of the ablated variants are defined as follows:

\begin{itemize}
    \item \textbf{w/o Router}: We replace our proposed Semantic-Aware Router with a standard MLP-based gating network. As analyzed in Section~\ref{app:complexity}, this substitution leads to an increase in trainable parameters due to the dense connections in the MLP layers.
    
    \item \textbf{w/o Scaling}: We disable the dynamic scaling mechanisms to verify their impact on task adaptation. Specifically, we fix all elements of the Diagonal Scaling Matrix $S$ to 1.0 and set the task-dependent scalar $g_{\text{task}}$ to 1.0 throughout the training process. Under this setting, the scaling strategy effectively reverts to the standard LoRA formulation.
\end{itemize}

\subsection{Analysis of Semantic-Aware Router}
\label{app:c2}

To strictly isolate the efficacy of our routing mechanism and eliminate interference from other components, we conduct a controlled experiment based on the asymmetric MoE-LoRA architecture (featuring one shared matrix $A$ and multiple semantic experts $B$).
In this setup, we vary only the routing module (comparing our Semantic-Aware Router against a standard MLP router) while keeping all other structures identical.
To make the expert specialization patterns more observable, we scale the number of experts to $N=8$ and employ Llama-3.1-8B as the backbone, training on the Commonsense Reasoning benchmark.

\paragraph{Expert Representation Analysis.}
As illustrated in Figure~\ref{fig:pca}, we visualize the Principal Component Analysis (PCA) projection of the learned expert features. 
The visualization reveals a stark contrast in the latent structure of the experts. 
With the MLP-based router, the expert representations tend to cluster closely together with ambiguous boundaries, indicating a high degree of functional overlap. 
In contrast, our SAMoRA framework produces highly distinct and separated expert clusters. 
This explicitly demonstrates that our approach successfully enforces expert distinctiveness, allowing each expert to specialize in different semantic subspaces.
\begin{figure*}[ht]
    \centering
    \includegraphics[width=\linewidth, height=5cm, keepaspectratio]{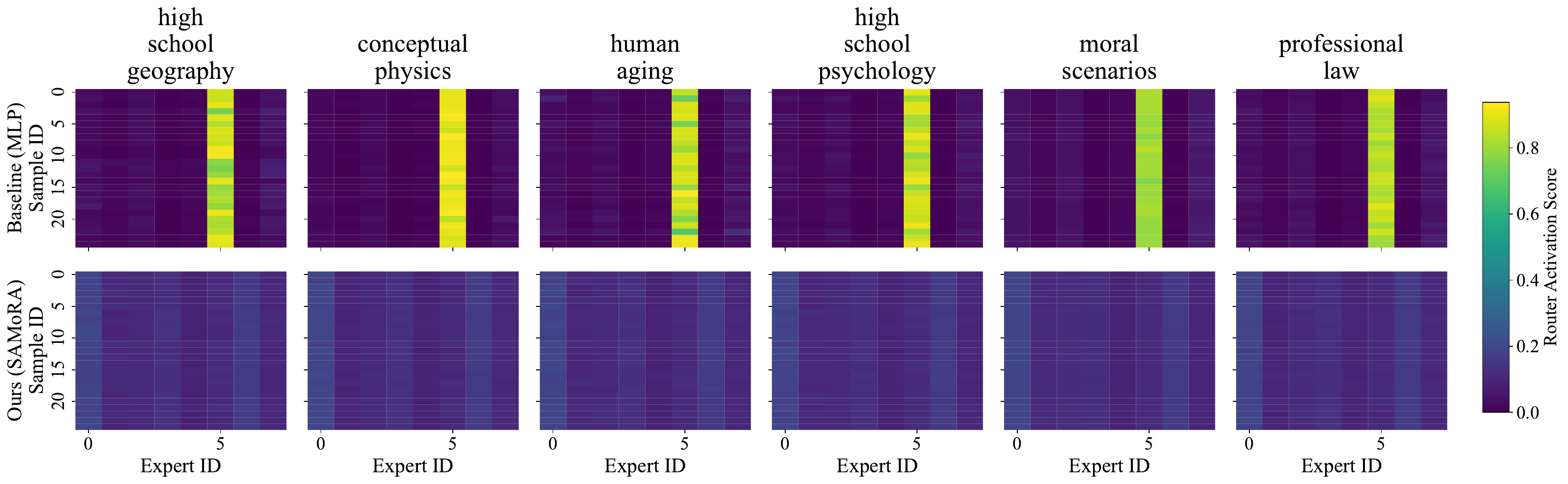}
    \caption{Visualization of expert activation patterns on the unseen MMLU benchmark. The top row (MLP Router) exhibits severe mode collapse, while our SAMoRA (bottom row) maintains diverse and adaptive routing across different subjects.}
    \label{fig:router_mmlu}
\end{figure*}
\paragraph{Routing Behavior on Unseen Tasks.}
To further evaluate the generalization capability of the router, we extend our analysis to the MMLU benchmark~\cite{DBLP:conf/nips/WangMZNCGRAHJLK24}, which serves as an unseen task during training. 
We visualize the proportion of activations for each expert in Figure~\ref{fig:router_mmlu}. 
Here, we display a subset of six randomly selected subjects characterized by diverse semantic distributions.
The figure is organized by subject columns, with the top row representing the MLP router and the bottom row representing ours.

A critical observation from the top row is that the MLP router suffers from severe representation collapse: regardless of the input subject, it predominantly selects \textbf{Expert 5}, with other experts being rarely activated. 
This behavior suggests that the MLP router fails to align expert specialization with input semantics, causing the dynamic MoE architecture to effectively degrade into a static, non-MoE model. 
Conversely, our method (bottom row) exhibits diverse and balanced activation patterns adaptive to different subjects, validating its ability to maintain precise routing even on out-of-distribution data.

\paragraph{Theoretical Rationale for Semantic Match Regularization.}
In our architecture, learnable Expert Keys represent the intended specialties. However, during unconstrained joint optimization, these keys risk diverging from the actual parameters the experts learn. We employ KL Divergence to penalize this structural misalignment. KL Divergence is theoretically optimal here because it rigorously measures the relative entropy between two probability distributions. By forcing the expert key's assignment distribution to closely track the intrinsic capability distribution (derived from expert weights), we steer the optimization trajectory toward a state where routing decisions are strictly anchored in the experts' genuine functional capabilities, rather than arbitrary local minima.

\paragraph{Theoretical Basis for Matrix $\mathbf{B}$ Row Averaging.}
\begin{itemize}
    \item \textbf{Isolating Expert Knowledge:} In our asymmetric LoRA structure ($\Delta W = \sum_{i=1}^{N} g_i B_i A$), the shared down-projection $A$ acts as a universal feature extractor, leaving matrix $B_i$ strictly responsible for the specialized mapping back to the output space. Consequently, $B_i$ inherently encodes the unique capabilities of that specific expert.
    \item \textbf{Geometric Centroid as Capability Anchor:} The rows of $B_i$ are transformation vectors residing in the $r$-dimensional latent space---the exact space where our routing occurs. By computing the average of these rows, we calculate the geometric centroid of the expert's parameter subspace. Theoretically, this centroid represents the dominant, macro-level semantic direction of the expert's transformations.
    \item \textbf{Robust and Dimensional Alignment:} This aggregation smooths out localized parameter noise, yielding a highly stable global representation in $\mathbb{R}^r$. This perfectly aligns with the dimensionality of the Expert Keys, allowing for a mathematically sound distance computation and ensuring the alignment loss is both meaningful and computationally efficient.
\end{itemize}
\begin{figure}[t]

    \centering

    \begin{subfigure}[b]{1.0\linewidth}

        \centering

        \includegraphics[width=\linewidth]{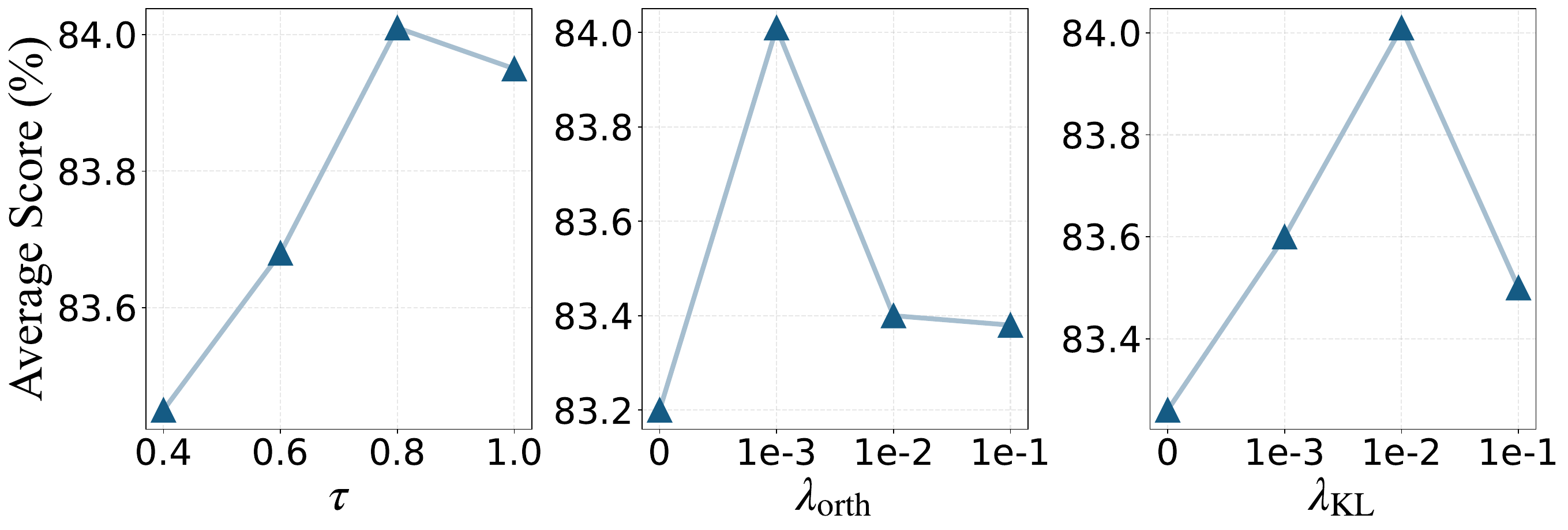}

        \caption{}

        \label{fig:sens_seq}

    \end{subfigure}


    \begin{subfigure}[b]{1.0\linewidth}

        \centering

        \includegraphics[width=\linewidth]{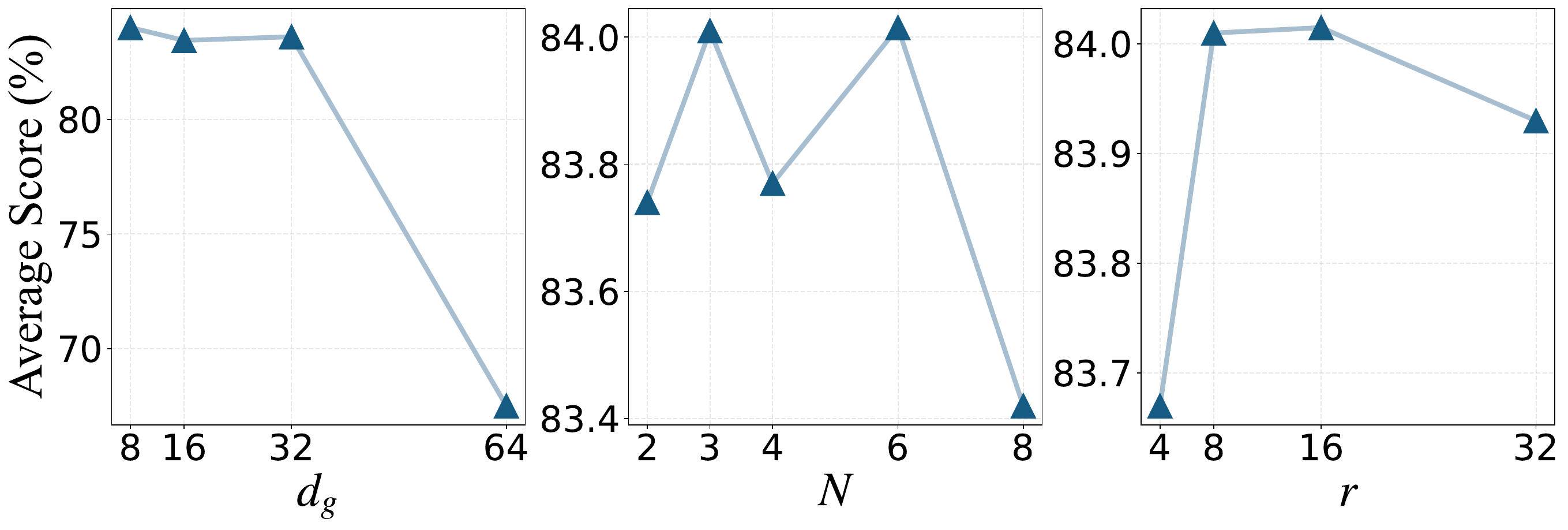}

        \caption{}

        \label{fig:sens_main}

    \end{subfigure}

    \caption{Sensitivity Analysis on hyperparameters evaluated on the Commonsense Reasoning dataset. Subfigure (a) and (b) illustrate different ablation settings.}

    \label{fig:combined_sens}

\end{figure}
\section{Hyperparameter Sensitivity}
\label{app:d}

We conduct a comprehensive sensitivity analysis to evaluate the robustness of our proposed framework under various hyperparameter configurations. 
All experiments in this section are performed on the Commonsense Reasoning benchmark using Llama-3.1-8B as the backbone model, trained for 1 epoch.
\subsection{Hyperparameter Sensitivity Analysis}

We conduct a comprehensive sensitivity analysis to investigate how different hyperparameter configurations affect SAMoRA's performance.

\paragraph{Impact of Model Architecture ($N, r, d_g$).}
We first evaluate the impact of model capacity ($N, r$) and the task embedding dimension ($d_g$) in Figure~\ref{fig:combined_sens}(b).
\begin{itemize}
    \item \textbf{Robustness to Capacity ($N, r$):} The performance remains relatively stable across a broad range of expert counts $N$ and LoRA ranks $r$. Specifically, increasing $r$ from 8 to 64 yields marginal gains, confirming that our method is parameter-efficient and does not rely on high-rank adapters. Similarly, the robustness against $N$ indicates that our routing mechanism effectively utilizes available experts without suffering from redundancy.
    \item \textbf{Task Embedding Dimension ($d_g$):} We observe a distinct behavior regarding $d_g$. While the model performs well with compact dimensions, there is a sharp accuracy drop when $d_g$ is increased to 64. This suggests that overly large task embeddings may introduce excessive parameters relative to the supervision signal, hindering convergence. Thus, a compact $d_g$ is sufficient for effective semantic encoding.
\end{itemize}

\paragraph{Impact of Optimization Hyperparameters ($\tau, \lambda_{\text{orth}}, \lambda_{\text{KL}}$).}
We further analyze the regularization terms and routing temperature.
Figure~\ref{fig:combined_sens}(a) illustrates the individual sensitivity trends for the temperature $\tau$, orthogonality loss weight $\lambda_{\text{orth}}$, and KL divergence weight $\lambda_{\text{KL}}$.
We observe that moderate values generally facilitate better convergence, preventing the router from collapsing or becoming too uniform.

To identify the optimal interaction between these terms, we report the joint ablation results in Table~\ref{tab:ab}.
\begin{itemize}
    \item \textbf{Temperature ($\tau$):} The temperature controls the sharpness of the routing distribution. We find that $\tau=0.8$ achieves the optimal performance (84.01\%). Lower temperatures (e.g., $\tau=0.4$) lead to premature expert collapse (83.45\%), while higher temperatures (e.g., $\tau=1.0$) result in an overly smooth distribution (83.95\%).
    \item \textbf{Regularization Weights:} Combined with $\tau=0.8$, appropriate regularization weights ($\lambda_{\text{orth}}$ and $\lambda_{\text{KL}}$) are essential to balance expert specialization and load distribution, securing the best trade-off between plasticity and stability.
\end{itemize}

\subsection{Impact of Loss Weights}
Finally, we analyze the sensitivity of the regularization hyperparameters: the orthogonality weight $\lambda_\text{orth}$ and the semantic match divergence weight $\lambda_\text{KL}$.

\paragraph{Orthogonality Weight ($\lambda_\text{orth}$).}
This term encourages diversity among experts. Comparing the rows in Table~\ref{tab:ab}:
\begin{itemize}
    \item Removing the regularization ($\lambda_\text{orth}=0$) results in a performance drop to 83.20\%, confirming the necessity of promoting expert diversity.
    \item However, setting $\lambda_\text{orth}$ too high (1E-2) causes a significant performance degradation to 79.35\%. This suggests that excessive constraints on orthogonality might hinder the optimization of the primary task loss.
    \item A moderate value of \textbf{1E-3} proves to be the most effective, striking a balance between expert diversity and task adaptation.
\end{itemize}

\paragraph{Semantic Match Weight ($\lambda_\text{KL}$).}
This term aligns the routing decisions with semantic information. The results show a positive correlation between $\lambda_\text{KL}$ and model performance within the tested range. Increasing $\lambda_\text{KL}$ from 0 to 1E-2 consistently improves accuracy (from 83.26\% to 84.01\%), highlighting the benefit of guiding the router with semantic knowledge derived from task embeddings.

\begin{table}[ht]

\centering

\begin{tabular}{lll|r}

\toprule

$\lambda_\text{orth}$ & $\lambda_\text{KL}$ & $\tau$& \textbf{Avg.}\\

\midrule

1E-3& 1E-2& 0.4 & 83.45\\

1E-3& 1E-2& 0.6 & 83.68\\

1E-3& 1E-2& 0.8 & \textbf{84.01}\\

1E-3& 1E-2& 1   & \underline{83.95}\\

1E-3& 1E-3& 0.8 & 83.61 \\

1E-3& 0 & 0.8 & 83.26 \\

1E-2& 1E-2 & 0.8 & 79.35 \\

0& 1E-2 & 0.8 & 83.20\\

\bottomrule

\end{tabular}
\caption{Sensitivity Analysis (\%) of regularization weights and temperature on Commonsense Reasoning dataset (Backbone: Llama-3.1-8B).}
\label{tab:ab}

\end{table}

\begin{table*}[t]
    \centering

    \begin{tabularx}{\linewidth}{l>{\raggedright\arraybackslash}X}
        \toprule
        \textbf{Task} & \textbf{Prompt Template} \\
        \midrule
        CoLA & Is the following sentence ``\{sentence\}'' grammatically acceptable? Answer: \\
        \addlinespace
        SST-2 & Is the following sentence ``\{sentence\}'' sentimently positive? Answer: \\
        \addlinespace
        MRPC & Does the following sentence ``\{sentence1\}'' convey the equivalent meaning as ``\{sentence2\}''? Answer: \\
        \addlinespace
        QQP & Is the following question ``\{question1\}'' essentially asking the same thing as ``\{question2\}''? Answer: \\
        \addlinespace
        MNLI & Does the statement ``\{premise\}'' imply that ``\{hypothesis\}''? Answer: \\
        \addlinespace
        QNLI & Based on the statement: ``\{question\}'' does the following sentence ``\{sentence\}'' have a definitive answer? Answer: \\
        \addlinespace
        RTE & Does the text ``\{sentence1\}'' entail the statement ``\{sentence2\}''? Answer: \\
        \bottomrule
    \end{tabularx}
    \caption{Prompt templates used for the Natural Language Understanding benchmark (GLUE). The placeholders (e.g., \{sentence\}) represent the input fields from the dataset.}
    \label{tab:nlu_prompts}
\end{table*}

\begin{table*}[t]
    \centering

    \resizebox{\textwidth}{!}{
    \begin{tabular}{l c c c c c c c}
        \toprule
        \textbf{Hyperparameter} & \textbf{LoRA} & \textbf{MultiLoRA} & \textbf{MoELoRA} & \textbf{HydraLoRA} & \textbf{MTL-LoRA} & \textbf{MoORE} & \textbf{SAMoRA}\\
        \midrule
        \multicolumn{8}{c}{\textit{Global Training Configurations}} \\
        \midrule
        Optimizer & \multicolumn{7}{c}{AdamW} \\
        Weight Decay & \multicolumn{7}{c}{0} \\
        $\beta_1$ & \multicolumn{7}{c}{0.9} \\
        $\beta_2$ & \multicolumn{7}{c}{0.95} \\
        Learning Rate & \multicolumn{7}{c}{$2\times10^{-4}$/ $3\times10^{-4}$} \\
        Batch Size & \multicolumn{7}{c}{8 / 64} \\
        Training Epochs & \multicolumn{7}{c}{3} \\
        Warmup Ratio & \multicolumn{7}{c}{0.01} \\
        Max Sequence Length & \multicolumn{7}{c}{512} \\
        Target Modules & \multicolumn{7}{c}{Q,K,V,O} \\
        \midrule
        \multicolumn{8}{c}{\textit{Method-Specific Architectures}} \\
        \midrule
        Rank ($r$) & 16 & 8 & 8 & 8 & 8 & 8 & 8\\
        Scale ($\alpha$) & 32 & 16 & 16 & 16 & 16 & 16 & -\\
        num\_experts ($N$) & - & 3 & 8 & 3 & 3 & 3 & 3\\
        Task Embedding Size ($d_g$) & - & - & 64 & - & - & 64 & 8\\
        Temperature ($\tau$) & - & - & - & - & 0.8 & - & 0.8\\
        $\lambda_\text{orth}$ & - & - & - & - & - & - & 1e-3 \\
        $\lambda_\text{match}$ & - & - & - & - & - & - & 1e-2 \\
        \bottomrule
    \end{tabular}
    }
    \caption{Detailed hyperparameter settings for all baseline methods on Commonsense Reasoning and GLUE benchmark. Common settings are listed in the top section, while method-specific parameters are detailed below. ``-'' indicates the parameter is not applicable.}
    \label{tab:hyperparameters}
\end{table*}
\end{document}